\documentclass[runningheads]{llncs}
\usepackage{graphicx}
\usepackage{comment}
\usepackage{amsmath,amssymb} 
\usepackage{color}
\usepackage{multirow}
\usepackage{amsmath}
\usepackage{array}

\begin{document}
\pagestyle{headings}
\mainmatter
\def\ECCVSubNumber{4652}  

\title{Joint Visual and Temporal Consistency for Unsupervised Domain Adaptive Person Re-Identification} 

\titlerunning{JVTC for Unsupervised Domain Adaptive Person ReID}

\author{Jianing Li, \and Shiliang Zhang}
\authorrunning{J. Li, and S. Zhang}
\institute{Department of Computer Science, School of EE\&CS, Peking University, \\Beijing 100871, China\\
\email{\{ljn-vmc,slzhang.jdl\}@pku.edu.cn}}

\maketitle

\begin{abstract}
Unsupervised domain adaptive person Re-IDentification (ReID) is challenging because of the large domain gap between source and target domains, as well as the lackage of labeled data on the target domain. This paper tackles this challenge through jointly enforcing visual and temporal consistency in the combination of a local one-hot classification and a global multi-class classification. The local one-hot classification assigns images in a training batch with different person IDs, then adopts a Self-Adaptive Classification (SAC) model to classify them.
The global multi-class classification is achieved by predicting labels on the entire unlabeled training set with the Memory-based Temporal-guided Cluster (MTC). MTC predicts multi-class labels by considering both visual similarity and temporal consistency to ensure the quality of label prediction.
The two classification models are combined in a unified framework, which effectively leverages the unlabeled data for discriminative feature learning. Experimental results on three large-scale ReID datasets demonstrate the superiority of proposed method in both unsupervised and unsupervised domain adaptive ReID tasks. For example, under unsupervised setting, our method outperforms recent unsupervised domain adaptive methods, which leverage more labels for training.

\keywords{Domain Adaption, Person Re-Identification, Convolution Neural Networks}
\end{abstract}

\section{Introduction}

Person Re-Identification (ReID) aims to identify a probe person in a camera network by matching his/her images or video sequences and has many promising applications like smart surveillance and criminal investigation. Recent years have witnessed the significant progresses on supervised person ReID in discriminative feature learning from labeled person images ~\cite{su2017pose,wei2017glad,li2019pose,li2018harmonious,zhong2020robust,mao2019resolution} and videos ~\cite{li2019global,li2019multi,li2020multi}. However, supervised person ReID methods rely on a large amount of labeled data which is expensive to annotate. Deep models trained on the source domain suffer substantial performance drop when transferred to a different target domain. Those issues make it hard to deploy supervised ReID models in real applications.

To tackle this problem, researchers focus on unsupervised learning~\cite{zhong2019invariance,fu2019self,wang2020unsupervised}, which could take advantage of abundant unlabeled data for training. Compared with supervised learning, unsupervised learning relieves the requirement for expensive data annotation, hence shows better potential to push person ReID towards real applications.
Recent works define unsupervised person ReID as a transfer learning task, which leverages labeled data on other domains. Related works can be summarized into two categories, \emph{e.g.},
(1) using Generative Adversarial Network (GAN) to transfer the image style from labeled source domain to unlabeled target domain while preserving identity labels for training~\cite{wei2018person,zhong2018camstyle,zhong2019invariance}, or
(2) pre-training a deep model on source domain, then clustering unlabeled data in target domain to estimate pseudo labels for training ~\cite{fu2019self,yu2019unsupervised}.
The second category has significantly boosted the performance of unsupervised person ReID. However, there is still a considerable performance gap between supervised and unsupervised person ReID. The reason may be because many persons share similar appearance and the same person could exhibit different appearances, leading to unreliable label estimation. Therefore, more effective ways to utilize the unlabeled data should still be investigated.

This work targets to learn discriminative features for unlabeled target domain through generating more reliable label predictions. Specifically, reliable labels can be predicted from two aspects. First, since each training batch samples a small number of images from the training set, it is likely that those images are sampled from different persons. We thus could label each image with a distinct person ID and separate them from each other with a classification model. Second, it is not reliable to estimate labels on the entire training set with only visual similarity. We thus consider both visual similarity and temporal consistency for multi-class label prediction, which is hence utilized to optimize the inter and intra class distances. Compared with previous methods, which only utilize visual similarity to cluster unlabeled images~\cite{fu2019self,yu2019unsupervised}, our method has potential to exhibit better robustness to visual variance. Our temporal consistency is inferred based on the video frame number, which can be easily acquired without requiring extra annotations or manual alignments.

The above intuitions lead to two classification tasks for feature learning.
The local classification in each training batch is conducted by a Self-Adaptive Classification (SAC) model. Specially, in each training batch, we generate a self-adaptive classifier from image features and apply one-hot label to separate images from each other. The feature optimization in the entire training set is formulated as a multi-label classification task for global optimization. We propose the Memory-based Temporal-Guided Cluster (MTC) to predict multi-class labels based on both visual similarity and temporal consistency. In other words, two images are assigned with the same label if they a) share large visual similarity and b) share enough temporal consistency.

Inspired by~\cite{wang2019spatial}, we compute the temporal consistency based on the distribution of time interval between two cameras, \emph{i.e.}, interval of frame numbers of two images. For example, when we observe a person appears in camera $i$ at time $t$, according to the estimated distribution, he/she would have high possibility to be recorded by camera $j$ at time $t+\Delta t$, and has low possibility will be recorded by another camera $k$. This cue would effectively filter hard negative samples with similar visual appearance, as well as could be applied in ReID to reduce the search space. To further ensure the accuracy of clustering result, MTC utilizes image features stored in the memory bank. Memory bank is updated with augmented features after each training iteration to improve feature robustness.


The two classification models are aggregated in a unified framework for discriminative feature learning. Experiments on three large-scale person ReID datasets show that, our method exhibits substantial superiority to existing unsupervised and domain adaptive ReID methods. For example, we achieve rank1 accuracy of 79.5\% on Market-1501 with unsupervised training, and achieve 86.8\% after unsupervised domain transfer, respectively.

Our promising performance is achieved with the following novel components. 1) The SAC model efficiently performs feature optimization in each local training batch by assigning images with different labels. 2) The MTC method performs feature optimization in the global training set by predicting labels with visual similarity and temporal consistency. 3) Our temporal consistency does not require any extra annotations or manual alignments, and could be utilized in both model training and ReID similarity computation. To the best of our knowledge, this is an early unsupervised person ReID work utilizing temporal consistency for label prediction and model training.

\section{Related Work}
This work is closely related to unsupervised domain adaptation and unsupervised domain adaptive person ReID. This section briefly summarizes those two categories of works.

\noindent\textbf{Unsupervised Domain Adaptation (UDA)} has been extensively studied in image classification. The aim of UDA is to align the domain distribution between source and target domains. A common solution of UDA is to define and minimize the domain discrepancy between source and target domain.
Gretton \emph{et.al.}~\cite{gretton2007kernel} project data samples into a reproducing kernel Hilbert space and compute the difference of sample means to reduce the Maximum Mean Discrepancy
(MMD).
Sun \emph{et.al.}~\cite{sun2016return} propose to learn a transformation to align the mean and covariance between two domains in the feature space.
Pan \emph{et.al.}~\cite{pan2019transferrable} propose to align each class in source and target domain through Prototypical Networks.
Adversarial learning is also widely used to minimize domain shift.
Ganin \emph{et.al.}~\cite{ganin2014unsupervised} propose a Gradient Reversal Layer (GRL) to confuse the feature learning model and make it can't distinguish the features from source and target domains.
DRCN~\cite{ghifary2016deep} takes a similar approach but also performs multi-task learning to reconstruct target domain images.
Different from domain adaption in person ReID, traditional UDA mostly assumes that the source domain and target domain share same classes. However, in person ReID, different domain commonly deals with different persons, thus have different classes.

\noindent\textbf{Unsupervised Domain Adaptive Person ReID:}
Early methods design hand craft features for person ReID ~\cite{gray2008viewpoint,liao2015person}. Those methods can be directly adapted to unlabeled dataset, but show unsatisfactory performance.
Recent works propose to train deep models on labeled source domain and then transfer to unlabeled target domain.
Yu \emph{et.al.}~\cite{yu2019unsupervised} use the labeled source dataset as a reference to learn soft labels.
Fu \emph{et.al.}~\cite{fu2019self} cluster the global and local features to estimate pseudo labels, respectively.
Generative Adversarial Network (GAN) is also applied to bridge the gap across cameras or domains.
Wei \emph{et.al.}~\cite{wei2018person} transfer images from the source domain to target domain while reserving the identity labels for training.
Zhong \emph{et.al.}~\cite{zhong2018camera} apply CycleGAN~\cite{zhu2017unpaired} to generate images under different camera styles for data augmentation.
Zhong \emph{et.al.}~\cite{zhong2019invariance} introduce the memory bank~\cite{wu2018unsupervised} to minimize the gap between source and target domains.

Most existing methods only consider visual similarity for feature learning on unlabeled data, thus are easily influenced by the large visual variation and domain bias. Different from those works, we consider visual similarity and temporal consistency for feature learning. Compared with existing unsupervised domain adaptive person ReID methods, our method exhibits stronger robustness and better performance. As shown in our experiments, our approach outperforms recent ReID methods under both unsupervised and unsupervised domain adaptive settings. To the best of our knowledge, this is an early attempt to jointly consider visual similarity and temporal consistency in unsupervised domain adaptive person ReID. Another person ReID work~\cite{wang2019spatial} also uses temporal cues. Different with our work, it focuses on supervised training and only uses temporal cues in the ReID stage for re-ranking.

\section{Proposed Method}

\subsection{Formulation}
For any query person image $q$, the person ReID model is expected to produce a feature vector to retrieve the image $g$ containing the same person from a gallery set. In other words, the ReID model should guarantee $q$ share more similar feature with $g$ than with other images. Therefore, learning a discriminative feature extractor is critical for person ReID.

In unsupervised domain adaptive person ReID, we have an unlabeled target domain $T = \{t_i\}_{i=1}^{N_T}$ containing $N_T$ person images. Additionally, a labeled source domain $ S = \{s_i, y_i\}_{i=1}^{N_S}$ containing $N_S$ labeled person images is provided as an auxiliary training set, where $y_i$ is the identity label associated with the person image $s_i$. The goal of domain adaptive person ReID is to learn a discriminative feature extractor $\operatorname f(\cdot)$ for $T$, using both $S$ and $T$.

The training of $\operatorname f(\cdot)$ can be conducted by minimizing the training loss on both source and target domains. With person ID labels, the training on $S$ can be considered as a classification task by minimizing the cross-entropy loss, \emph{i.e.},
\begin{equation}
\begin{aligned}
\mathcal{L}_{src}=-\frac{1}{N_S}\sum_{i=1}^{N_S}\operatorname {log}\ \operatorname P(y_i|s_i),
\end{aligned}
\label{equ:softmax}
\end{equation}
where $\operatorname P(y_i|s_i)$ is the predicted probability of sample $s_i$ belonging to class $y_i$. This supervised learning ensures the performance of $\operatorname f(\cdot)$ on source domain.

To gain discriminative power of $\operatorname f(\cdot)$ to the target domain, we further compute training loss with predicted labels on $T$. First, because each training batch samples $n_T, n_T\ll N_T$ images from $T$, it is likely that $n_T$ images are sampled from different persons. We thus simply label each image $t_i$ in the mini-batch with a distinct person ID label, \emph{i.e.}, an one-hot vector $\boldsymbol l_i$ with $\boldsymbol{l}_i[j]=1$ only if $i=j$. A Self-Adaptive Classification (SAC) model is adopted to separate images of different persons in the training batch. The objective of SAC can be formulated as minimizing the classification loss, \emph{i.e.},
\begin{equation}
\begin{aligned}
\mathcal{L}_{local} = \frac{1}{n_T}\sum_{i=1}^{n_T} \operatorname {L}(\mathcal{V} \times \operatorname f(t_i), \boldsymbol l_i),
\end{aligned}
\label{equ:loss1}
\end{equation}
where $n_T$ denotes the number of images in a training batch. $\operatorname f(\cdot)$ produces a $d$-dim feature vector. $\mathcal{V}$ stores $n_T$ $d$-dim vectors as the classifier. $\mathcal{V} \times \operatorname f(t_i)$ computes the classification score, and $\operatorname {L}(\cdot)$ computes the loss by comparing classification scores and one-hot labels. Details of classifier $\mathcal{V}$ will be given in Sec.~\ref{sec:sac}.

Besides the local optimization in each training batch, we further predict labels on the entire $T$ and perform a global optimization. Since each person may have multiple images in $T$, we propose the Memory-based Temporal-guide Cluster (MTC) to predict a multi-class label for each image.
For an image $t_i$, MTC predicts its multi-class label $\boldsymbol m_i$, where $\boldsymbol m_i[j]=1$ only if $t_i$ and $t_j$ are regarded as containing the same person.

Predicted label $\boldsymbol m_i$ allows for a multi-label classification on $T$. We introduce a memory bank $\mathcal{K}\in \mathbf R^{N_T\times d}$ to store $N_T$ image features as a $N_T$-class classifier~\cite{zhong2019invariance}.
The multi-label classification loss is computed by classifying image feature $\operatorname f(t_i)$ with the memory bank $\mathcal{K}$, then comparing the classification scores with multi-class label $\boldsymbol m_i$. The multi-label classification loss on $T$ can be represented as
\begin{equation}
\begin{aligned}
\mathcal{L}_{global} = \frac{1}{N_T}\sum_{i=1}^{N_T} \operatorname {L}(\mathcal{K}\times \operatorname f(t_i), \boldsymbol m_i),
\end{aligned}
\label{equ:global}
\end{equation}
where $\mathcal{K}\times \operatorname f(t_i)$ produces the classification score. The memory bank $\mathcal{K}$ is updated after each training iteration as
\begin{equation}
\begin{aligned}
\mathcal{K}[i]^t = (1-\alpha) \mathcal{K}[i]^{t-1} + \alpha \operatorname f(t_i),
\end{aligned}
\label{equ:bank2}
\end{equation}
where the superscript $t$ denotes the training epoch, $\alpha$ is the updating rate. Detailed of MTC and $\boldsymbol m_i$ computation will be presented in Sec.~\ref{sec:mtc}.

By combining the above losses computed on $S$ and $T$, the overall training loss of our method can be formulated as,
\begin{equation}
\begin{aligned}
\mathcal{L} = \mathcal{L}_{src} + w_1\mathcal{L}_{local} + w_2\mathcal{L}_{global},
\end{aligned}
\label{equ:loss}
\end{equation}
where $w_1$ and $w_2$ are loss weights.

The accuracy of predicted labels, \emph{i.e.}, $\boldsymbol l$ and $\boldsymbol m$ is critical for the training on $T$. The accuracy of $\boldsymbol l$ can be guaranteed by setting batch size $n_T\ll N_T$, and using careful sampling strategies. To ensure the accuracy of $\boldsymbol m$, MTC considers both visual similarity and temporal consistency for label prediction.

\begin{figure}[t]
\centering
\includegraphics[width=0.98\linewidth]{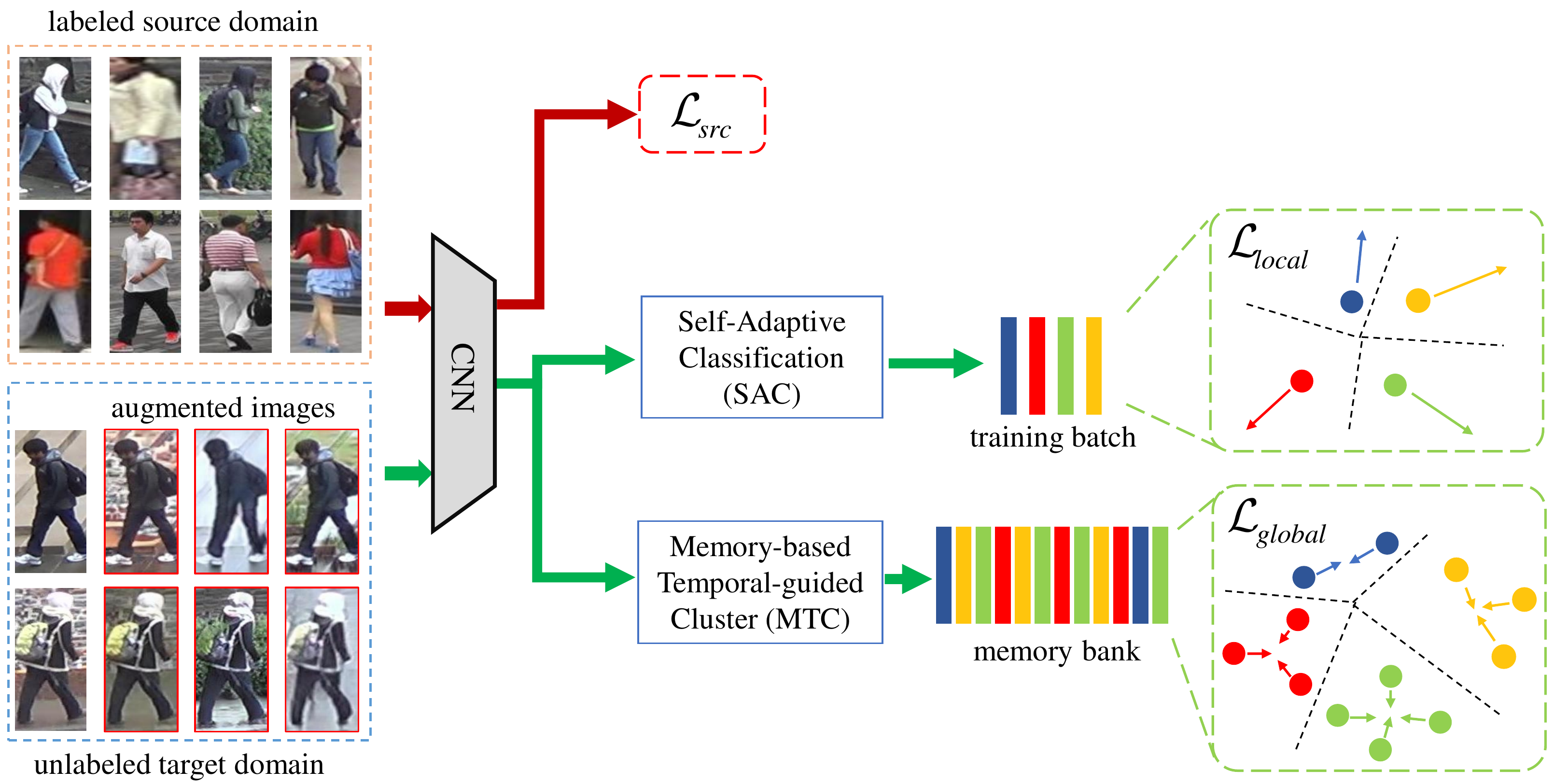}\\
\caption{Overview of the proposed framework for unsupervised domain adaptive ReID model training. $\mathcal{L}_{src}$ is computed on the source domain. SAC computes $\mathcal{L}_{local}$ in each training batch. MTC computes $\mathcal{L}_{global}$ on the entire target domain. SAC and MTC predict one-hot label and multi-class label for each image, respectively. Without $\mathcal{L}_{src}$, our framework works as unsupervised training.}
\label{fig:framework}
\end{figure}

We illustrate our training framework in Fig.~\ref{fig:framework}, where $\mathcal{L}_{local}$ can be efficiently computed within each training batch by classifying a few images. $\mathcal{L}_{global}$ is a more powerful supervision by considering the entire training set $T$. The combination of $\mathcal{L}_{local}$ and $\mathcal{L}_{global}$ utilizes both temporal and visual consistency among unlabeled data and guarantees strong robustness of the learned feature extractor $\operatorname f(\cdot)$. The following parts proceed to introduces the computation of $\mathcal{L}_{local}$ in SAC, and $\mathcal{L}_{local}$ in MTC, respectively.

\subsection{Self-Adaptive Classification}\label{sec:sac}

SAC classifies unlabeled data in each training batch. As shown in Eq.~\eqref{equ:loss1}, the key of SAC is the classifier $\mathcal V$. For a batch consisting of $n_T$ images, the classifier $\mathcal{V}$ is defined as a $n_T\times d$ sized tensor, where the $i$-th $d$-dim vector represents the classifiers for the $i$-th image. To enhance its robustness, $\mathcal{V}$ is calculated based on features of original images and their augmented duplicates.

Specifically, for an image $t_i$ in training batch, we generate $k$ images $t_{i}^{(j)}\ (j=1,2,...,k)$ with image argumentation. This enlarges the training batch to $n_T\times (k+1)$ images belonging to $n_T$ categories. The classifier $\mathcal{V}$ is computed as,
\begin{equation}
\begin{aligned}
\mathcal{V}=[v_1, v_2,...v_{n_T}]\in \mathbf R^{n_T\times d},\ v_i=\frac{1}{k+1}{(\operatorname f(t_i)+\sum_{j=1}^k \operatorname f(t_{i}^{(j)}))},
\end{aligned}
\label{equ:sac0}
\end{equation}
where $v_i$ is the averaged feature of $t_i$ and its augmented images. It can be inferred that, the robustness of $\mathcal V$ enhances as $\operatorname f(\cdot)$ gains more discriminative power. We thus call $\mathcal V$ as a self-adapted classifier.

Data augmentation is critical to ensure the robustness of $\mathcal V$ to visual variations. We consider each camera as a style domain and adopt CycleGAN~\cite{zhu2017unpaired} to train camera style transfer models~\cite{zhong2018camera}. For each image under a specific camera, we totally generate $C-1$ images with different styles, where $C$ is the camera number in the target domain. We set $k<C-1$. Therefore, each training batch randomly selects $k$ augmented images for training.

Based on classifier $\mathcal{V}$ and the one-hot label $\boldsymbol l$, the $\mathcal{L}_{local}$ of SAC can be formulated as the cross-entropy loss, \emph{i.e.},
\begin{equation}
\mathcal{L}_{local} = -\frac{1}{n_T\times (k+1)}\sum _{i=1}^{n_T} (\operatorname {log}( \operatorname P(i| t_{i}) + \sum _{j=1}^k \operatorname{log}(\operatorname P(i| t_{i}^{(j)})),
\label{equ:sac2}
\end{equation}
where $\operatorname P(i| t_{i})$ is the probability of image $t_{i}$ being classified to label $i$, \emph{i.e.},
\begin{equation}
\begin{aligned}
\operatorname P(i| t_{i})=\frac{\operatorname {exp}(v_{i}^T \cdot \operatorname f(t_i)/\beta_1)}{\sum_{n=1}^{n_T} \operatorname {exp}(v_n^T \cdot \operatorname f(t_i)/\beta_1)}
\end{aligned}
\label{equ:sac1}
\end{equation}
where $\beta_1$ is a temperature factor to balance the feature distribution.

$\mathcal{L}_{local}$ can be efficiently computed on $n_T$ images. Minimizing $\mathcal{L}_{local}$ enlarges the feature distance of images in the same training batch, meanwhile decreases the feature distance of augmented images in the same category. It thus boosts the discriminative power of $\operatorname f(\cdot)$ on $T$.

\subsection{Memory-based Temporal-guided Cluster}\label{sec:mtc}

\begin{figure}[t]
\centering
\includegraphics[width=0.95\linewidth]{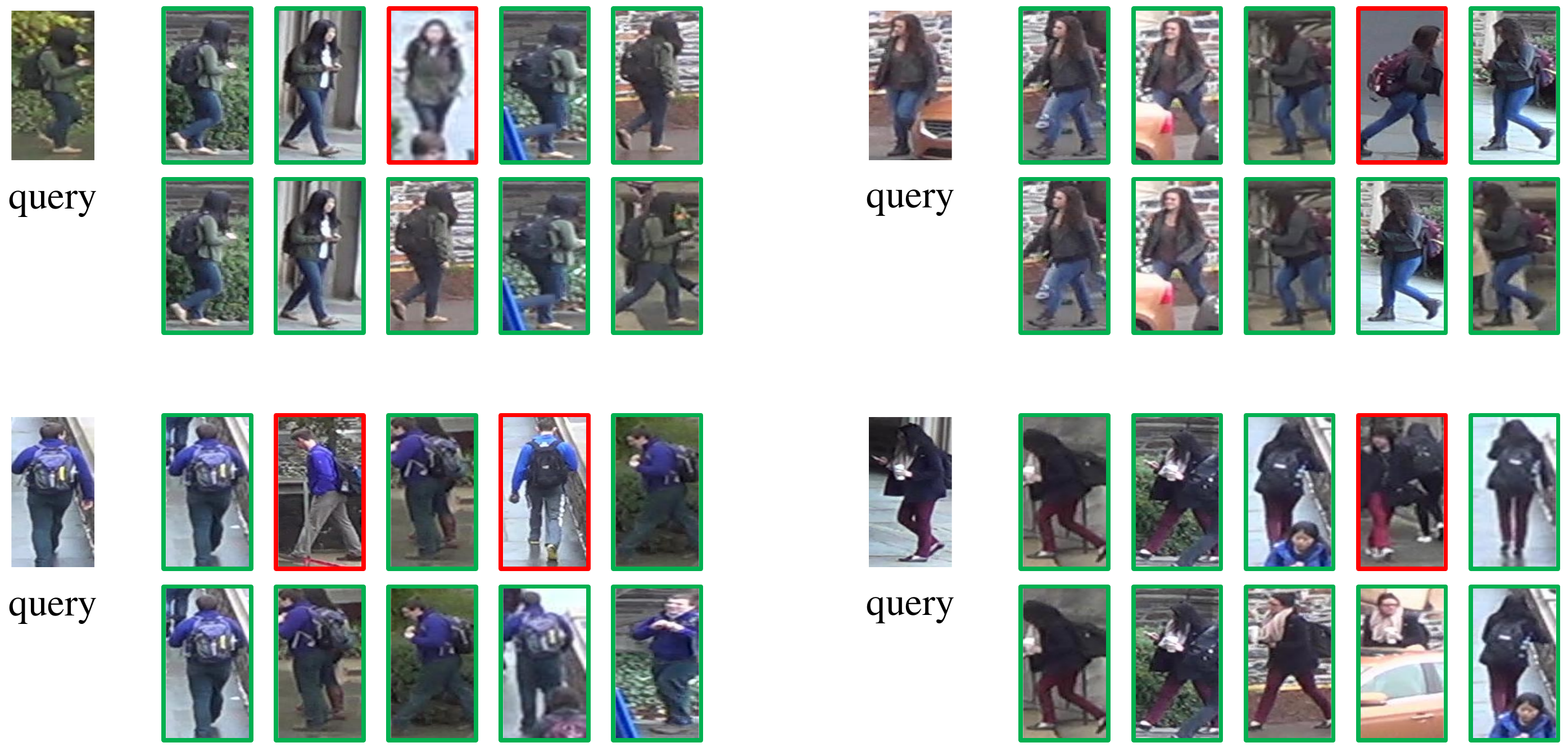}\\
\caption{Illustration of person ReID results on DukeMTMC-reID dataset. Each example shows the top-5 retrieved images by visual similarity (first tow) and joint similarity computed in Eq.~\eqref{equ:dist} (second row). The true match is annotated by the green bounding box and false match is annotated by the red bounding box.}
\label{fig:temporal}
\end{figure}

MTC predicts the multi-class label $\boldsymbol m_i$ for image $t_i$ through clustering images in $T$, \emph{i.e.}, images inside the same cluster are assigned with the same label. The clustering is conducted based on the pair-wise similarity considering both visual similarity and temporal consistency of two images.

Visual similarity can be directly computed using the feature extractor $\operatorname f(\cdot)$ or the features stored in the memory bank $\mathcal{K}$. Using $\operatorname f(\cdot)$ requires to extract features for each image in $T$, which introduces extra time consumption. Meanwhile, the features in $\mathcal{K}$ can be enhanced by different image argumentation strategies, making them more robust. We hence use features in $\mathcal{K}$ to compute the visual similarity between two images $t_i$ and $t_j$, \emph{i.e.},
\begin{equation} \label{eq:vsim}
\operatorname {vs}(t_i,t_j) = \operatorname {cosine}(\mathcal{K}[i], \mathcal{K}[j]),
\end{equation}
where $\operatorname {vs}(\cdot)$ computes the visual similarity with cosine distance.

Temporal consistency is independent to visual features and is related to the camera id and frame id of each person image. Suppose we have two images $t_i$ from camera $a$ and $t_j$ from camera $b$ with frame IDs $fid_i$ and $fid_j$, respectively. The temporal consistency between $t_i$ and $t_j$ can be computed as,
\begin{equation}
\begin{aligned}
\operatorname{ts}(t_i,t_j)=\operatorname{{H}}_{(a,b)}(fid_i-fid_j),
\end{aligned}
\label{equ:tc3}
\end{equation}
where $\operatorname{{H}}_{(a,b)}(\cdot)$ is a function for camera pair ($a$, $b$). It estimates the temporal consistency based on frame id interval of $t_i$ and $t_j$, which reflects the time interval when they are recorded by cameras $a$ and $b$.

$\operatorname{{H}}_{(a,b)}(\cdot)$ can be estimated based on a histogram $\operatorname {\bar H}_{(a,b)}(int)$, which shows the probability of appearing identical person at camera $a$ and $b$ for frame id interval $int$. $\operatorname {\bar H}_{(a,b)}(int)$ can be easily computed on datasets with person ID labels. To estimate it on unlabeled $T$, we first cluster images in $T$ with visual similarity in Eq.~\eqref{eq:vsim} to acquire pseudo person ID labels. Suppose $n_{(a,b)}$ is the total number of image pairs containing identical person in camera $a$ and $b$. The value of $int$-th bin in histogram, \emph{i.e.}, $\operatorname {\bar H}_{(a,b)}(int)$ is computed as,
\begin{equation}
\begin{aligned}
\operatorname {\bar H}_{(a,b)}(int) = {n^{int}_{(a,b)}}/{n_{(a,b)}},
\end{aligned}
\label{equ:tc1}
\end{equation}
where $n^{int}_{(a,b)}$ is the number of image pairs containing identical person in camera $a$ and $b$ with frame id intervals $int$.

For a dataset with $C$ cameras, ${C(C-1)}/{2}$ histograms will be computed. We finally use Gaussian function to smooth the histogram and take the smoothed histogram as ${\operatorname{H}}_{(a,b)}(\cdot)$ for temporal consistency computation.

Our final pair-wise similarity is computed based on $\operatorname {vs}(\cdot)$ and $\operatorname {ts}(\cdot)$. Because those two similarities have different value ranges, we first normalize them, then perform the fusion. This leads to the joint similarity function $\operatorname J(\cdot)$, \emph{i.e.},
\begin{equation}
\begin{aligned}
\operatorname J(t_i,t_j) = {1}/{(1+\lambda _0 e^{-\gamma _0 \operatorname{vs}(t_i,t_j)})} \times {1}/{(1+\lambda _1 e^{-\gamma _1 \operatorname{ts}(t_i,t_j)})},
\end{aligned}
\label{equ:dist}
\end{equation}
where $\lambda_0$ and $\lambda_1$ are smoothing factors, $\gamma_0$ and $\gamma_1$ are shrinking factors.

Eq.~\eqref{equ:dist} computes more reliable similarities between images than either Eq.~\eqref{eq:vsim} or Eq.~\eqref{equ:tc3}. $\operatorname J(\cdot)$ can also be used in person ReID for query-gallery similarity computation. Fig.~\ref{fig:temporal} compares some ReID results achieved by visual similarity and joint similarity, respectively. It can be observed that, the joint similarity is more discriminative than the visual similarity.

We hence cluster images in target domain $T$ based on $\operatorname J(\cdot)$ and assign the multi-class label for each image. For an image $t_i$, its multi-class label $\boldsymbol m_i[j]= 1$ only if $t_i$ and $t_j$ are in the same cluster. Based on $\boldsymbol m$, the $\mathcal{L}_{global}$ on $T$ can be computed as,
\begin{equation}
\begin{aligned}
\mathcal{L}_{global} = - \frac{1}{N_T}\sum_{i=1}^{N_T}\sum_{j=1}^{N_T} \boldsymbol m_i[j]\times \operatorname {log} \operatorname {\bar P}(j|t_i)/|\boldsymbol m_i|_1,
\end{aligned}
\label{equ:global}
\end{equation}
where $|\cdot|_1$ computes the L-1 norm. $\operatorname {\bar P}(j|t_i)$ denotes the probability of image $t_i$ being classified to the $j$-th class in multi-label classification, \emph{i.e.},
\begin{equation}
\begin{aligned}
\operatorname {\bar P}(j|t_i)=\frac{\operatorname {exp}(\mathcal{K}[j]^T \cdot \operatorname f(t_i)/\beta_2)}{\sum_{n=1}^{N_T} \operatorname {exp}(\mathcal{K}[n]^T \cdot \operatorname f(t_i)/\beta_2)},
\end{aligned}
\label{equ:bank1}
\end{equation}
where $\beta_2$ is the temperature factor. The following section proceeds to discuss the effects of parameters and conduct comparisons with recent works.

\section{Experiment}
\subsection{Dataset}
We evaluate our methods on three widely used person ReID datasets, \emph{e.g.}, Market1501~\cite{zheng2015scalable}, DukeMTMC-ReID~\cite{zheng2017unlabeled,ristani2016performance}, and MSMT17~\cite{wei2018person}, respectively.

\textbf{Market1501} consists of 32,668 images of 1,501 identities under 6 cameras. The dataset is divided into training and test sets, which contains 12,936 images of 751 identities and 19,732 images of 750 identities, respectively.

\textbf{DukeMTMC-ReID} is composed of 1,812 identities and 36,411 images under 8 cameras. 16,522 images of 702 pedestrians are used for training. The other identities and images are included in the testing set.

\textbf{MSMT17} is currently the largest image person ReID dataset. MSMT17 contains 126,441 images of 4,101 identities under 15 cameras. The training set of MSMT17 contains 32,621 bounding boxes of 1,041 identities, and the testing set contains 93,820 bounding boxes of 3,060 identities.

We follow the standard settings in previous works~\cite{fu2019self,zhong2019invariance} for training in domain adaptive person ReID and unsupervised person ReID, respectively. Performance is evaluated by the Cumulative Matching Characteristic (CMC) and mean Average Precision (mAP). We use \textbf{JVTC} to denote our method.

\subsection{Implementation Details}
We adopt ResNet50~\cite{he2016deep} as the backbone and add a $512$-dim embedding layer for feature extraction. We initialize the backbone with the model pre-trained on ImageNet~\cite{deng2009imagenet}. All models are trained and finetuned with PyTorch. Stochastic Gradient Descent (SGD) is used to optimize our model. Input images are resized to $256\times 128$. The mean value is subtracted from each (B, G, and R) channel. The batch size is set as $128$ for both source and target domains. Each training batch in the target domain contains $32$ original images and each image has 3 augmented duplicates, \emph{i.e.}, we set $k$=3.

The temperature factor $\beta _1$ is set as $0.1$ and $\beta _2$ is set as $0.05$. The smoothing factors and shrinking factors $\lambda _0$, $\lambda _1$, $\gamma _0$ and $\gamma _1$ in Eq.~\eqref{equ:dist} are set as $1,2,5$ and $5$, respectively. The initial learning rate is set as $0.01$, and is reduced by ten times after $40$ epoches. The multi-class label $\boldsymbol m$ are updated every 5 epochs based on visual similarity initially, and the joint similarity is introduced at $30$-th epoch. Only local loss $\mathcal{L}_{local}$ is applied at the initial epoch. The $\mathcal{L}_{global}$ is applied at the $10$-th epoch. The training is finished after $100$ epoches. The memory updating rate $\alpha$ starts from 0 and grows linearly to 1. The loss weights $w_1$ and $w_2$ are set as $1$ and $0.2$, respectively. DBSCAN~\cite{ester1996density} is applied for clustering.

\subsection{Ablation Study}

\begin{table}[t]
\caption{Evaluation of individual components of JVTC.}
\label{table:ablation}
\setlength{\tabcolsep}{3 pt}
\footnotesize
\begin{center}
\begin{tabular}{l|c|c|c|c|c|c|c|c|c|c}
\hline
Dataset    &\multicolumn{5}{c|}{DukeMTMC $\rightarrow$ Market1501}&\multicolumn{5}{c}{Market1501 $\rightarrow$ DukeMTMC}\\
\hline
Method             &mAP  &r1&r5&r10&r20  &mAP  &r1&r5&r10&r20\\
\hline
Supervised   &69.7&86.3&94.3&96.5&97.6  &61.0&80.2&89.1&91.9&94.2\\
Direct transfer  &18.2&42.1&60.7&67.9&74.8  &16.6&31.8&48.4&55.0&61.7\\
\hline
Baseline      &46.6&77.4&89.5&93.0&95.1  &43.6&66.1&77.7&81.7&84.8\\
SAC           &41.8&64.5&76.0&79.6&92.3  &37.5&59.4&74.1&78.3&81.4\\
MTC           &56.4&79.8&91.0&93.9&95.9  &51.1&71.3&81.1&84.3&86.3\\
JVTC          &61.1&83.8&93.0&95.2&96.9  &56.2&75.0&85.1&88.2&90.4\\
JVTC+         &\textbf{67.2}&\textbf{86.8}&\textbf{95.2}&\textbf{97.1}&\textbf{98.1} &\textbf{66.5}&\textbf{80.4}&\textbf{89.9}&\textbf{92.2}&\textbf{93.7}\\
\hline
\end{tabular}
\end{center}
\end{table}

\subsubsection{Evaluation of Individual Components:}
This section investigates the effectiveness of each component in our framework, \emph{e.g.}, the SAC and MTC. We summarize the experimental results in Table~\ref{table:ablation}. In the table, ``Supervised" denotes training deep models with labeled data on the target domain, and testing on the testing set. ``Direct transfer" denotes directly using the model trained on source domain for testing. ``Baseline" uses memory bank for multi-label classification, but predicts multi-class label only based on visual similarity.
``SAC" is implemented based on ``Direct transfer" by applying SAC model for one-hot classification. ``MTC" utilizes both visual similarity and temporal consistency for multi-class label prediction. ``JVTC" combines SAC and MTC. ``JVTC+" denotes using the joint similarity for person ReID.

Table~\ref{table:ablation} shows that, supervised learning on the target domain achieves promising performance. However, directly transferring the supervised model to different domains leads to substantial performance drop, \emph{e.g.}, the rank1 accuracy drops to $44.2\%$ on Market1501 and $48.4\%$ on DukeMTMC-reID after direct transfer. The performance drop is mainly caused by the domain bias between datasets.

It is also clear that, SAC consistently outperforms direct transfer by large margins. For instance, SAC improves the rank1 accuracy from $42.1\%$ to $64.5\%$ and $31.8\%$ to $59.4\%$ on Market-1501 and DukeMTMC-reID, respectively. This shows that, although SAC is efficient to compute, it effectively boosts the ReID performance on target domain. Compared with the baseline, MTC uses joint similarity for label prediction. Table~\ref{table:ablation} shows that, MTC performs better than the baseline, \emph{e.g.}, outperforms baseline by 9.8\% and 5.2\% in mAP on Market1501 and DukeMTMC-reID, respectively.
This performance gain clearly indicates the robustness of our joint similarity.

After combining SAC and MTC, JVTC achieves more substantial performance gains on two datasets. For instance, JVTC achieves mAP of 61.1\% on Market1501, much better than the 46.6\% of baseline. ``JVTC+" further uses joint similarity to compute the query-gallery similarity. It achieves the best performance, and outperforms the supervised training on target domain. We hence could conclude that, each component in our method is important for performance boost, and their combination achieves the best performance.

\begin{figure}[t]
\centering
\includegraphics[width=0.95\linewidth]{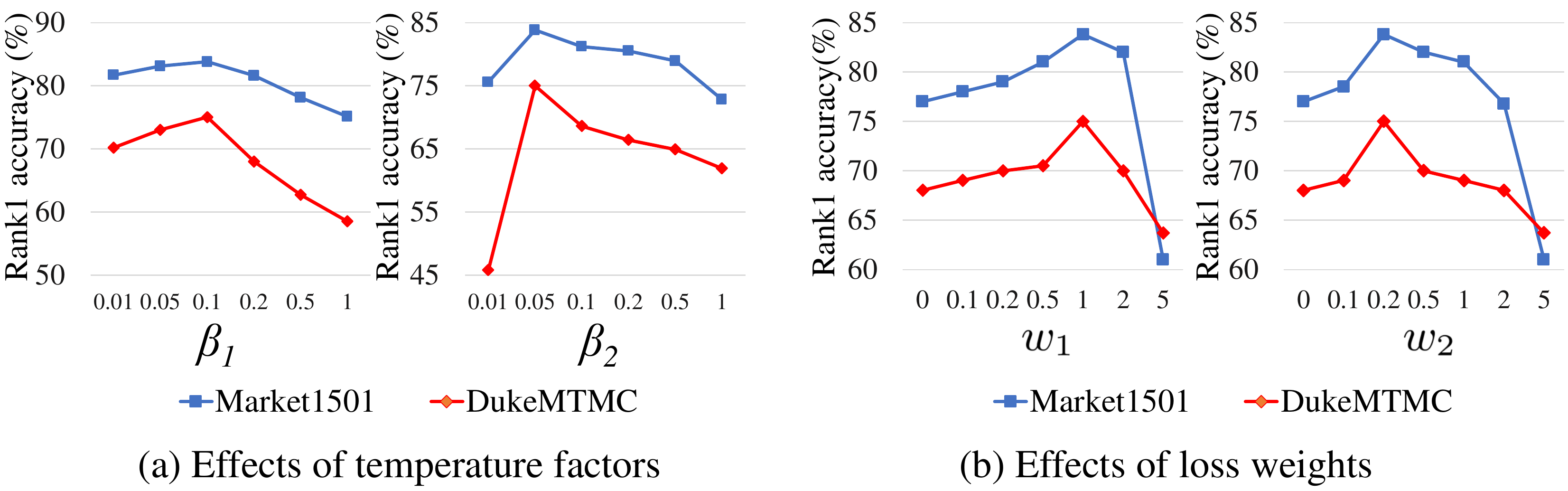}\\
\caption{Influences of temperature factors $\beta_1$ and $\beta_2$ in (a), and loss weights $w_1$, $w_2$ in (b). Experiments are conducted on Market1501 and DukeMTMC-reID.}
\label{fig:weight}
\end{figure}

\noindent
\textbf{Hyper-parameter Analysis:}
This section investigates some important hyper-parameters in our method, including the temperature factors $\beta _1$, $\beta _2$, and the loss weights $w _1$ and $w _2$, respectively. To make the evaluation possible, each experiment varies the value of one hyper-parameter while keeping others fixed. All experiments are conducted with unsupervised domain adaptive ReID setting on both Market-1501 and DukeMTMC-reID.

Fig.~\eqref{fig:weight}(a) shows the effects of temperature factors $\beta _1$ and $\beta _2$ in Eq.~\eqref{equ:sac1} and Eq.\eqref{equ:bank1}. We can see that, a small temperature factor usually leads to better ReID performance. This is because that smaller temperature factor leads to a smaller entropy in the classification score, which is commonly beneficial for classification loss computation. However, too small temperature factor makes the training hard to converge. According to Fig.~\ref{fig:weight}(a), we set $\beta _1=0.1$, $\beta _2=0.05$.

Fig.~\ref{fig:weight}(b) shows effects of loss weight $w_1$ and $w_2$ in network training.
We vary the loss weight $w_1$ and $w_2$ from $0$ to $5$.
$w_1(w_2)=0$ means we don't consider the corresponding loss during training. It is clear that, a positive loss weight is beneficial for the ReID performance on both datasets. As we increase the loss weights, the ReID performance starts to increase. The best performance is achieved with $w_1=1$ and $w_2=0.2$ on two datasets. Further increasing the loss weights substantially drops the the ReID performance. This is because increasing $w_1$ and $w_2$ decreases the weight of $\mathcal{L}_{src}$, which is still important. Based on this observation, we set $w_1=1$ and $w_2=0.2$ in following experiments.

\begin{table}[t]
\caption{Comparison with unsupervised, domain adaptive, and semi-supervised ReID methods on Market1501 and DukeMTMC-reID.}
\label{table:comparison}
\setlength{\tabcolsep}{1.5 pt}
\footnotesize
\begin{center}
\resizebox{1\linewidth}{!}{
\begin{tabular}{l|cccccc|cccccc}
\hline
Dataset    &\multicolumn{6}{c|}{Market1501}&\multicolumn{6}{c}{DukeMTMC}\\
\hline
Method       &Source&mAP  &r1&r5&r10&r20  &Source&mAP &r1&r5&r10&r20\\
\hline
Supervised  &Market&69.7&86.3&94.3&96.5&97.6  &Duke&61.0&80.2&89.1&91.9&94.2\\
Direct transfer   &Duke  &18.2&42.1&60.7&67.9&74.8  &Market&16.6&31.8&48.4&55.0&61.7\\
\hline
LOMO~\cite{liao2015person}      &None&8.0 &27.2&41.6&49.1&-   &None&4.8 &12.3&21.3&26.6&-\\
BOW~\cite{zheng2015scalable}    &None&14.8&35.8&52.4&60.3&-   &None&8.3 &17.1&28.8&34.9&-\\
BUC~\cite{lin2019bottom}        &None&38.3&66.2&79.6&84.5&-   &None&27.5&47.4&62.6&68.4&-\\
DBC~\cite{ding12dispersion}     &None&41.3&69.2&83.0&87.8&-   &None&30.0&51.5&64.6&70.1&-\\
JVTC                            &None &{41.8}&{72.9}&{84.2}&{88.7}&{92.0} &None&{42.2}&{67.6}&{78.0}&{81.6}&{84.5}\\
JVTC+                           &None&\textbf{47.5}&\textbf{79.5}&\textbf{89.2}&\textbf{91.9}&\textbf{94.0}  &None&\textbf{50.7}&\textbf{74.6}&\textbf{82.9}&\textbf{85.3}&\textbf{87.2}\\
\hline\hline
PTGAN~\cite{wei2018person}         &Duke &-   &38.6&-&66.1&-       &Market&-&27.4&-&50.7&-\\
CamStyle~\cite{zhong2018camstyle}  &Duke &27.4&58.8&78.2&84.3&88.8 &Market&25.1&48.4&62.5&68.9&74.4\\
T-Fusion~\cite{lv2018unsupervised} &CUHK01&-  &60.8&74.4&79.3&-    &-&-&-&-&-&-\\
ARN~\cite{li2018adaptation}        &Duke &39.4&70.3&80.4&86.3&93.6 &Market&33.4&60.2&73.9&79.5&82.5\\
MAR~\cite{yu2019unsupervised}      &MSMT17&40.0&67.7&81.9&87.3&-   &MSMT17&48.0&67.1&79.8&84.2&-\\
ECN~\cite{zhong2019invariance}     &Duke &43.0&75.1&87.6&91.6&-    &Market&40.4&63.3&75.8&80.4&-\\
PDA-Net~\cite{li2019cross}         &Duke &47.6&75.2&86.3&90.2&-    &Market&45.1&63.2&77.0&82.5&-\\
PAST ~\cite{zhang2019self}         &Duke &54.6&78.4&-&-&-          &Market&54.3&72.4&-&-&-\\
CAL-CCE~\cite{qi2019novel}         &Duke &49.6&73.7&-&-&-          &Market&45.6&64.0&-&-&-\\
CR-GAN~\cite{chen2019instance}     &Duke&54.0&77.7&89.7&92.7&-     &Market&48.6&68.9&80.2&84.7&-\\
SSG~\cite{fu2019self}              &Duke&58.3&80.0&90.0&92.4&-     &Market&53.4&73.0&80.6&83.2&-\\
\hline
TAUDL~\cite{li2018unsupervised} &Tracklet&41.2&63.7&-&-&- &Tracklet&43.5&61.7&-&-&-\\
UTAL~\cite{li2019unsupervised}  &Tracklet&46.2&69.2&-&-&- &Tracklet&43.5&62.3&-&-&-\\
SSG+~\cite{fu2019self}             &Duke&62.5&81.4&91.6&93.8&-     &Market&56.7&74.2&83.5&86.7&-\\
SSG++~\cite{fu2019self}            &Duke&\textbf{68.7}&86.2&94.6&96.5&-     &Market&60.3&76.0&85.8&89.3&-\\
\hline
JVTC      &Duke&{61.1}&{83.8}&{93.0}&{95.2}&{96.9} &Market&{56.2}&{75.0}&{85.1}&{88.2}&{90.4}\\
JVTC+     &Duke&{67.2}&\textbf{86.8}&\textbf{95.2}&\textbf{97.1}&\textbf{98.1} &Market&\textbf{66.5}&\textbf{80.4}&\textbf{89.9}&\textbf{92.2}&\textbf{93.7}\\
\hline
\end{tabular}}
\end{center}
\end{table}

\subsection{Comparison with State-of-the-art Methods}
\label{sec:comparison}
This section compares our method against state-of-the-art unsupervised, unsupervised domain adaptive, and semi-supervised methods on three datasets. Comparisons on Market1501 and DukeMTMC-reID are summarized in Table~\ref{table:comparison}. Comparisons on MSMT17 are summarized in Table~\ref{table:msmt}. In those tables, ``Source" refers to the labeled source dataset, which is used for training in unsupervised domain adaptive ReID. ``None" denotes unsupervised ReID.

\textbf{Comparison on Market1501 and DukeMTMC-reID:}
We first compare our method with unsupervised learning methods.
Compared methods include hand-crafted features LOMO~\cite{liao2015person} and BOW~\cite{zheng2015scalable}, and deep learning methods DBC~\cite{ding12dispersion} and BUC~\cite{lin2019bottom}.  It can be observed from Table~\ref{table:comparison} that,
hand-crafted features LOMO and BOW show unsatisfactory performance, even worse than directly transfer. Using unlabeled training dataset for training, deep learning based methods outperform hand-crafted features. BUC and DBC first treat each image as a single cluster, then merge clusters to seek pseudo labels for training. Our method outperforms them by large margins, \emph{e.g.}, our rank1 accuracy on Market1501 achieves 72.9\% \emph{vs.} their 66.2\% and 69.2\%, respectively. The reasons could be because our method considers both visual similarity and temporal consistency to predict labels. Moreover, our method further computes classification loss in each training batch with SAC. By further considering temporal consistency during testing, JVTC+ gets further performance promotions on both datasets, even outperforms several unsupervised domain adaptive methods.

We further compare our method with unsupervised domain adaptive methods including PTGAN~\cite{wei2018person}, CamStyle~\cite{zhong2018camstyle},  T-Fusion~\cite{lv2018unsupervised},
ARN~\cite{li2018adaptation}, MAR~\cite{yu2019unsupervised}, ECN~\cite{zhong2019invariance}, PDA-Net~\cite{li2019cross}, PAST ~\cite{zhang2019self}, CAL-CCE~\cite{qi2019novel}, CR-GAN~\cite{chen2019instance} and SSG~\cite{fu2019self}, and semi-supervised methods including TAUDL~\cite{li2018unsupervised}, UTAL~\cite{li2019unsupervised}, SSG+~\cite{fu2019self}, and SSG++~\cite{fu2019self}.
Under the unsupervised domain adaptive training setting, our method achieves the best performance on both Market1501 and DukeMTMC-reID in Table~\ref{table:comparison}. For example, our method achieves 83.8\% rank1 accuracy on Market1501 and gets 75.0\% rank1 accuracy on DukeMTMC-reID.
T-Fusion~\cite{lv2018unsupervised} also use temporal cues for unsupervised ReID, but achieves unsatisfactory performance, \emph{e.g.}, 60.8\% rank1 accuracy on Market1501 dataset. The reason may because that T-Fusion directly multiplies the visual and temporal probabilities, while our method fuses the visual and temporal similarities through more reasonable smooth fusion to boost the robustness.
Our method also consistently outperforms the recent SSG~\cite{fu2019self} on those two datasets. SSG clusters multiple visual features and needs to train 2100 epoches before convergence. Differently, our method only uses global feature and could be well-trained in 100 epoches. We hence could conclude that, our method is also more efficient than SSG. By further considering temporal consistency during testing, JVTC+ outperforms semi-supervised method SSG++~\cite{fu2019self} and supervised training on target domain.

\begin{table}[t]
\caption{Comparison with unsupervised and domain adaptive methods on MSMT17.}
\label{table:msmt}
\footnotesize
\begin{center}
\begin{tabular}{l|c|c|c|c|c|c}
\hline
Method   &Source                &mAP &r1&r5&r10&r20 \\
\hline
Supervised &MSMT17 &35.9&63.3&77.7&82.4&85.9 \\
\hline
JVTC       &None &15.1&39.0&50.9&56.8&61.9\\
JVTC+      &None &17.3&43.1&53.8&59.4&64.7\\
\hline
PTGAN~\cite{wei2018person}    &\multirow{6}{*}{Market1501} &2.9 &10.2 &24.4 &-  \\
ECN~\cite{zhong2019invariance}&&8.5 &25.3&36.3&42.1&-\\
SSG~\cite{fu2019self}         &&13.2&31.6&49.6&-&-\\
SSG++~\cite{fu2019self}       &&16.6&37.6&57.2&-&-\\
JVTC                          &&19.0&42.1&53.4&58.9&64.3\\
JVTC+                         &&\textbf{25.1}&\textbf{48.6}&\textbf{65.3}&\textbf{68.2}&\textbf{75.2}\\
\hline
PTGAN~\cite{wei2018person}    &\multirow{6}{*}{DukeMTMC}&3.3 &11.8&27.4&-&-\\
ECN~\cite{zhong2019invariance}&&10.2&30.2&41.5&46.8&-\\
SSG~\cite{fu2019self}         &&13.3&32.2&51.2&-&-\\
SSG++~\cite{fu2019self}       &&18.3&41.6&62.2&-&-\\
JVTC                          &&20.3&45.4&58.4&64.3&69.7\\
JVTC+                         &&\textbf{27.5}&\textbf{52.9}&\textbf{70.5}&\textbf{75.9}&\textbf{81.2}\\
\hline
\end{tabular}
\end{center}
\end{table}

\textbf{Comparison on MSMT17:} MSMT17 is more challenging than Market1501 and DukeMTMC-reID because of more complex lighting and scene variations. Some works have reported performance on MSMT17, including unsupervised domain adaptive methods PTGAN~\cite{wei2018person}, ECN~\cite{zhong2019invariance} and SSG~\cite{fu2019self}, and semi-supervised method SSG++~\cite{fu2019self}, respectively.
The comparison on MSMT17 are summarized in Table~\ref{table:msmt}.
As shown in the table, our method outperforms existing methods by large margins. For example, our method achieves 45.4\% rank1 accuracy when using DukeMTMC-reID as the source dataset, which outperforms the unsupervised domain adaptive method SSG~\cite{fu2019self} and semi-supervised method SSG++~\cite{fu2019self} by 13.2\% and 3.8\%, respectively. We further achieves 52.9\% rank1 accuracy after applying the joint similarity during ReID. This outperforms the semi-supervised method SSG++~\cite{fu2019self} by 11.3\%. The above experiments on three datasets demonstrate the promising performance of our JVTC.

\section{Conclusion}
This paper tackles unsupervised domain adaptive person ReID through jointly enforcing visual and temporal consistency in the combination of local one-hot classification and global multi-class classification.
Those two classification tasks are implemented by SAC and MTC, respectively. SAC assigns images in the training batch with distinct person ID labels, then adopts a self-adaptive classier to classify them. MTC predicts multi-class labels by considering both visual similarity and temporal consistency to ensure the quality of label prediction. The two classification models are combined in a unified framework for discriminative feature learning on target domain. Experimental results on three datasets demonstrate the superiority of the proposed method over state-of-the-art unsupervised and domain adaptive ReID methods.

~\\
\begin{small}
\textbf{Acknowledgments}
This work is supported in part by Peng Cheng Laboratory, The National Key Research and Development Program of China under Grant No. 2018YFE0118400, in part by Beijing Natural Science Foundation under Grant No. JQ18012, in part by Natural Science Foundation of China under Grant No. 61936011, 61620106009, 61425025, 61572050, 91538111.
\end{small}
\newpage

\bibliographystyle{splncs04}
\bibliography{egbib}

\begin{thebibliography}{10}
\providecommand{\url}[1]{\texttt{#1}}
\providecommand{\urlprefix}{URL }
\providecommand{\doi}[1]{https://doi.org/#1}

\bibitem{chen2019instance}
Chen, Y., Zhu, X., Gong, S.: Instance-guided context rendering for cross-domain
  person re-identification. In: ICCV (2019)

\bibitem{deng2009imagenet}
Deng, J., Dong, W., Socher, R., Li, L.J., Li, K., Fei-Fei, L.: Imagenet: A
  large-scale hierarchical image database. In: CVPR (2009)

\bibitem{ding12dispersion}
Ding12, G., Khan, S., Yin12, Q., Tang12, Z.: Dispersion based clustering for
  unsupervised person re-identification. In: BMVC (2019)

\bibitem{ester1996density}
Ester, M., Kriegel, H.P., Sander, J., Xu, X.: Density-based spatial clustering
  of applications with noise. In: KDD (1996)

\bibitem{fu2019self}
Fu, Y., Wei, Y., Wang, G., Zhou, Y., Shi, H., Huang, T.S.: Self-similarity
  grouping: A simple unsupervised cross domain adaptation approach for person
  re-identification. In: ICCV (2019)

\bibitem{ganin2014unsupervised}
Ganin, Y., Lempitsky, V.: Unsupervised domain adaptation by backpropagation.
  arXiv preprint arXiv:1409.7495  (2014)

\bibitem{ghifary2016deep}
Ghifary, M., Kleijn, W.B., Zhang, M., Balduzzi, D., Li, W.: Deep
  reconstruction-classification networks for unsupervised domain adaptation.
  In: ECCV (2016)

\bibitem{gray2008viewpoint}
Gray, D., Tao, H.: Viewpoint invariant pedestrian recognition with an ensemble
  of localized features. In: ECCV (2008)

\bibitem{gretton2007kernel}
Gretton, A., Borgwardt, K., Rasch, M., Sch{\"o}lkopf, B., Smola, A.J.: A kernel
  method for the two-sample-problem. In: NeurIPS (2007)

\bibitem{he2016deep}
He, K., Zhang, X., Ren, S., Sun, J.: Deep residual learning for image
  recognition. In: CVPR (2016)

\bibitem{li2019global}
Li, J., Wang, J., Tian, Q., Gao, W., Zhang, S.: Global-local temporal
  representations for video person re-identification. In: ICCV (2019)

\bibitem{li2019multi}
Li, J., Zhang, S., Huang, T.: Multi-scale 3d convolution network for video
  based person re-identification. In: AAAI (2019)

\bibitem{li2020multi}
Li, J., Zhang, S., Huang, T.: Multi-scale temporal cues learning for video
  person re-identification. IEEE Trans. on Image Processing  \textbf{29},
  4461--4473 (2020)

\bibitem{li2019pose}
Li, J., Zhang, S., Tian, Q., Wang, M., Gao, W.: Pose-guided representation
  learning for person re-identification. IEEE Trans. on Pattern Analysis and
  Machine Intelligence  (2019)

\bibitem{li2018unsupervised}
Li, M., Zhu, X., Gong, S.: Unsupervised person re-identification by deep
  learning tracklet association. In: ECCV (2018)

\bibitem{li2019unsupervised}
Li, M., Zhu, X., Gong, S.: Unsupervised tracklet person re-identification. IEEE
  Trans. on Pattern Analysis and Machine Intelligence  (2019)

\bibitem{li2018harmonious}
Li, W., Zhu, X., Gong, S.: Harmonious attention network for person
  re-identification. In: CVPR (2018)

\bibitem{li2019cross}
Li, Y.J., Lin, C.S., Lin, Y.B., Wang, Y.C.F.: Cross-dataset person
  re-identification via unsupervised pose disentanglement and adaptation. arXiv
  preprint arXiv:1909.09675  (2019)

\bibitem{li2018adaptation}
Li, Y.J., Yang, F.E., Liu, Y.C., Yeh, Y.Y., Du, X., Frank~Wang, Y.C.:
  Adaptation and re-identification network: An unsupervised deep transfer
  learning approach to person re-identification. In: CVPR Workshops (2018)

\bibitem{liao2015person}
Liao, S., Hu, Y., Zhu, X., Li, S.Z.: Person re-identification by local maximal
  occurrence representation and metric learning. In: CVPR (2015)

\bibitem{lin2019bottom}
Lin, Y., Dong, X., Zheng, L., Yan, Y., Yang, Y.: A bottom-up clustering
  approach to unsupervised person re-identification. In: AAAI (2019)

\bibitem{lv2018unsupervised}
Lv, J., Chen, W., Li, Q., Yang, C.: Unsupervised cross-dataset person
  re-identification by transfer learning of spatial-temporal patterns. In: CVPR
  (2018)

\bibitem{mao2019resolution}
Mao, S., Zhang, S., Yang, M.: Resolution-invariant person re-identification.
  In: IJCAI (2019)

\bibitem{pan2019transferrable}
Pan, Y., Yao, T., Li, Y., Wang, Y., Ngo, C.W., Mei, T.: Transferrable
  prototypical networks for unsupervised domain adaptation. In: CVPR (2019)

\bibitem{qi2019novel}
Qi, L., Wang, L., Huo, J., Zhou, L., Shi, Y., Gao, Y.: A novel unsupervised
  camera-aware domain adaptation framework for person re-identification. arXiv
  preprint arXiv:1904.03425  (2019)

\bibitem{ristani2016performance}
Ristani, E., Solera, F., Zou, R., Cucchiara, R., Tomasi, C.: Performance
  measures and a data set for multi-target, multi-camera tracking. In: ECCV
  (2016)

\bibitem{su2017pose}
Su, C., Li, J., Zhang, S., Xing, J., Gao, W., Tian, Q.: Pose-driven deep
  convolutional model for person re-identification. In: ICCV (2017)

\bibitem{sun2016return}
Sun, B., Feng, J., Saenko, K.: Return of frustratingly easy domain adaptation.
  In: AAAI (2016)

\bibitem{wang2020unsupervised}
Wang, D., Zhang, S.: Unsupervised person re-identification via multi-label
  classification. In: CVPR (2020)

\bibitem{wang2019spatial}
Wang, G., Lai, J., Huang, P., Xie, X.: Spatial-temporal person
  re-identification. In: AAAI (2019)

\bibitem{wei2018person}
Wei, L., Zhang, S., Gao, W., Tian, Q.: Person transfer gan to bridge domain gap
  for person re-identification. In: CVPR (2018)

\bibitem{wei2017glad}
Wei, L., Zhang, S., Yao, H., Gao, W., Tian, Q.: Glad: Global-local-alignment
  descriptor for pedestrian retrieval. In: ACM MM (2017)

\bibitem{wu2018unsupervised}
Wu, Z., Xiong, Y., Yu, S.X., Lin, D.: Unsupervised feature learning via
  non-parametric instance discrimination. In: CVPR (2018)

\bibitem{yu2019unsupervised}
Yu, H.X., Zheng, W.S., Wu, A., Guo, X., Gong, S., Lai, J.H.: Unsupervised
  person re-identification by soft multilabel learning. In: CVPR (2019)

\bibitem{zhang2019self}
Zhang, X., Cao, J., Shen, C., You, M.: Self-training with progressive
  augmentation for unsupervised cross-domain person re-identification. In: ICCV
  (2019)

\bibitem{zheng2015scalable}
Zheng, L., Shen, L., Tian, L., Wang, S., Wang, J., Tian, Q.: Scalable person
  re-identification: A benchmark. In: ICCV (2015)

\bibitem{zheng2017unlabeled}
Zheng, Z., Zheng, L., Yang, Y.: Unlabeled samples generated by gan improve the
  person re-identification baseline in vitro. In: ICCV (2017)

\bibitem{zhong2020robust}
Zhong, Y., Wang, X., Zhang, S.: Robust partial matching for person search in
  the wild. In: CVPR (2020)

\bibitem{zhong2019invariance}
Zhong, Z., Zheng, L., Luo, Z., Li, S., Yang, Y.: Invariance matters: Exemplar
  memory for domain adaptive person re-identification. In: CVPR (2019)

\bibitem{zhong2018camera}
Zhong, Z., Zheng, L., Zheng, Z., Li, S., Yang, Y.: Camera style adaptation for
  person re-identification. In: CVPR (2018)

\bibitem{zhong2018camstyle}
Zhong, Z., Zheng, L., Zheng, Z., Li, S., Yang, Y.: Camstyle: A novel data
  augmentation method for person re-identification. IEEE Trans. on Image
  Processing  \textbf{28}(3),  1176--1190 (2018)

\bibitem{zhu2017unpaired}
Zhu, J.Y., Park, T., Isola, P., Efros, A.A.: Unpaired image-to-image
  translation using cycle-consistent adversarial networks. In: ICCV (2017)

\end{thebibliography}
\end{document}